Original Paper

# Analyzing Geospatial and Socioeconomic Disparities in Breast Cancer Screening Among Populations in the United States: Machine Learning Approach

Soheil Hashtarkhani[1], PhD; Yiwang Zhou[2], PhD; Fekede Asefa Kumsa[1], PhD; Shelley White-Means[3], PhD; David L Schwartz[4], MD; Arash Shaban-Nejad[1], MPH, PhD

[1]Center for Biomedical Informatics, Department of Pediatrics, College of Medicine, University of Tennessee Health Science Center, Memphis, TN, United States

[2]Department of Biostatistics, St. Jude Children's Research Hospital, Memphis, TN, United States

[3]College of Graduate Health Sciences, University of Tennessee Health Science Center, Memphis, TN, United States

[4]Department of Radiation Oncology, College of Medicine, University of Tennessee Health Science Center, Memphis, TN, United States

**Corresponding Author:**

Arash Shaban-Nejad, MPH, PhD
Center for Biomedical Informatics, Department of Pediatrics, College of Medicine
University of Tennessee Health Science Center
50 N. Dunlap St
Memphis, TN, 38103
United States
Phone: 1 9012875836
Email: ashabann@uthsc.edu

## Abstract

**Background:** Breast cancer screening plays a pivotal role in early detection and subsequent effective management of the disease, impacting patient outcomes and survival rates.

**Objective:** This study aims to assess breast cancer screening rates nationwide in the United States and investigate the impact of social determinants of health on these screening rates.

**Methods:** Data on mammography screening at the census tract level for 2018 and 2020 were collected from the Behavioral Risk Factor Surveillance System. We developed a large-scale dataset of social determinants of health, comprising 13 variables for 72,337 census tracts. Spatial analysis employing Getis-Ord Gi statistics was used to identify clusters of high and low breast cancer screening rates. To evaluate the influence of these social determinants, we implemented a random forest model, with the aim of comparing its performance to linear regression and support vector machine models. The models were evaluated using $R^2$ and root mean squared error metrics. Shapley Additive Explanations values were subsequently used to assess the significance of variables and direction of their influence.

**Results:** Geospatial analysis revealed elevated screening rates in the eastern and northern United States, while central and midwestern regions exhibited lower rates. The random forest model demonstrated superior performance, with an $R^2$=64.53 and root mean squared error of 2.06, compared to linear regression and support vector machine models. Shapley Additive Explanations values indicated that the percentage of the Black population, the number of mammography facilities within a 10-mile radius, and the percentage of the population with at least a bachelor's degree were the most influential variables, all positively associated with mammography screening rates.

**Conclusions:** These findings underscore the significance of social determinants and the accessibility of mammography services in explaining the variability of breast cancer screening rates in the United States, emphasizing the need for targeted policy interventions in areas with relatively lower screening rates.

## Introduction

In the United States, breast cancer ranks as the second most prevalent form of cancer among women, surpassed only by skin cancer [1]. Annually, approximately 240,000 cases of breast cancer are diagnosed in women, and tragically, approximately 42,000 women succumb to this disease each year in the United States. This makes breast cancer the second leading cause of cancer-related mortality among women in the country, following lung cancer [2]. Screening for breast cancer serves as a crucial secondary prevention measure, aimed at identifying the disease at an early stage, prior to clinical manifestation. Early detection of breast cancer enables the implementation of less intensive treatment strategies, contributing to improved survival rates. Mammography-based screening detects lesions before they achieved clinical visibility [3]. Evidence shows that high-quality routine screening programs have led to a 25% to 31% reduction in breast cancer–related mortality among women aged 50 to 69 years [4].

The US Preventive Services Task Force recommends mammography every two years for women aged 40 to 74 years [5]. Despite these recommendations, current research indicates disparities in mammography screening across different parameters, including variations among women residing in different regions and belonging to different races, with varying levels of median household incomes, health insurance statuses, and access to mammography services [6]. The 2022 Cancer Trends Progress Report revealed that 76% of women aged 50-74 years underwent mammogram testing, with rates varying from 74% among Hispanic women to 82% among non-Hispanic Black women [7]. Additionally, 64% of women with less than a high school education, 67.5% of women with incomes below 200% of the federal poverty level, and 75% of those who were Medicare beneficiaries underwent a mammogram test [7]. The Healthy People 2030 [8] has set a target to increase the proportion of breast cancer screenings to 80% [9].

Geospatial and machine learning models have proven effective in identifying the impact of social, natural, and built environments on health outcomes [10-12]. This study seeks to explain the geographical disparities in breast cancer screening across the United States and to explore the area-level socioeconomic factors associated with the rates of breast cancer screening. By examining these disparities, we seek to provide insights that can guide targeted interventions and policies aimed at improving equitable access to breast cancer screening services.

## Methods

This cross-sectional study investigates the spatial and socioeconomic factors influencing mammography screening rates among women aged 50 to 74 years in the United States. The methodology section outlines the steps, including data collection, variable selection, descriptive analysis, spatial analysis, machine learning model implementation, and model performance evaluation.

### Data Collection

#### Dependent Variables

The data for mammography screening rates in this study were sourced from the Centers for Disease Control and Prevention's (CDC) PLACES Project for the years 2018 and 2020, which used responses collected through the Behavioral Risk Factors Surveillance System (BRFSS) survey [13]. This survey specifically targeted female respondents aged 50-74 years, categorizing them as women who reported having undergone mammogram screenings and those who did not (excluding unknowns and refusals). Our data extraction process comprised two main stages. First, we extracted age-adjusted mammography screening rates at the county level for spatial analysis, facilitating the visualization of patterns across the entire country. Second, we obtained the crude rates (raw percentages) of mammography screening at the census tract level, a small geographic unit used by the US Census Bureau for collecting and analyzing statistical data, explanatory analysis, and prediction model development by machine learning methods.

#### Independent Variables

Based on a preliminary literature review, we selected independent variables for the study from various sources. We incorporated socioeconomic data from the CDC, the 2013-2017 American Community Survey, the United States Department of Agriculture, and the Health Resources and Services Administration. The analyzed variables encompass a range of factors, including urban-rural location, population density, the rate of older women (aged 55 to 74 years), poverty rate, ethnicity (Black and Hispanic), educational attainment, uninsured rate, median home value, social vulnerability index, and primary care shortage area.

To assess accessibility, we used data from the US Food and Drug Administration's Mammography Facility Database, which included geocoding the locations of 8706 mammography centers. The geodesic distance from each census tract to the nearest facility and the number of facilities within a 10-mile radius were calculated. Table 1 provides a comprehensive overview of the dependent and independent variables used in this study, including their names, sources, and definitions.

**Table 1.** Dependent and independent variables used in this study.

| Variable name | Source | Unit | Definition |
|---|---|---|---|
| **Dependent variables** | | | |
| Mammography rate (2018) | CDC[a] | Percent | Crude percent of mammography use among women aged 50-74 years in 2018 |
| Mammography rate (2020) | CDC | Percent | Crude percent of mammography use among women aged 50-74 years in 2020 |
| **Independent variables** | | | |
| Urban-rural location | USDA[b] | Binary | Urban or rural tract as of 2019 |
| Population density | 2013-2017 ACS[c] | Per square mile | Number of people per square mile |
| Number of women aged ≥55 years | 2013-2017 ACS | Percent | Estimated percent of the female population aged 55 or above |
| Poverty rate | Census ACS data | Percent | Estimated percent of all people that are living in poverty |
| Without health insurance | 2013-2017 ACS | Percent | Estimated percent of the population without health insurance coverage |
| Higher education rate | 2013-2017 ACS | Percent | Estimated percent of the population ≥25 years, with a bachelor's, graduate, or professional degree |
| Black population | 2013-2017 ACS | Percent | Percent of the population that is Black or African American, by single census classification |
| Hispanic population | 2013-2017 ACS | Percent | Percent of the population identified as Hispanic or Latino |
| Home value | 2013-2017 ACS | Dollar | Estimated median value of an owner-occupied housing unit |
| Social vulnerability index | CDC | Index | Social vulnerability level as of 2020 |
| Primary care shortage | HRSA[d] | Binary | Primary care health professional shortage area status as of 2020 |
| Distance to nearest mammography facility | Calculated | Mile | Distance from the center of the census tract to the nearest accredited mammography facility |
| Number of mammography facilities | Calculated | Number | Number of mammography facilities within the 10-mile catchment of the census tract |

[a]CDC: Center for Disease Control and Prevention.
[b]USDA: United States Department of Agriculture.
[c]ACS: American Community Survey.
[d]HRSA: Health Resources and Services Administration.

## Analysis

### Preprocessing

The primary objective of preprocessing was to handle missing values of both dependent and independent variables within the dataset. Due to the complexity of accounting for both spatial and temporal correlations in imputing breast cancer screening rates, we opted to exclude any census tracts that lacked mammography screening data in the BRFSS dataset for the years 2018 and 2020. Missing independent variables were imputed using the mean values for numerical data and the mode for binary data from the 20 closest neighboring records.

### Thematic Mapping and Spatial Clustering

The age-adjusted rates for breast cancer screening were integrated into a shapefile of ArcGIS containing 3143 counties across all 50 states and the District of Columbia. Subsequently, the data was visualized using the natural break method [14] to enhance clarity. Using the Getis-Ord Gi statistic [15], we identified hotspots indicating areas with either high or low mammography screening rates. This spatial analysis allowed us to discern localized patterns and trends of breast cancer screening behavior.

### Machine Learning Analysis

While constructing the predictive model, the response variable was the mean value of mammography screening rates in 2018 and 2020 for each census tract. The dataset was randomly split into two parts: 75% was used for training the model, and the remaining 25% was reserved for testing. This division allowed us to develop the model using the training data and then assess its predictive performance on the unseen testing data.

In this study, an ensemble learning algorithm known as random forest (RF) was employed to model the relationship between geospatial factors and breast cancer screening rates. Ensemble learning combines multiple models to improve the overall prediction accuracy and robustness, which is the rationale for choosing RF [16].

To enhance the efficacy of the RF model, we conducted a systematic hyperparameter search, where a predefined grid of values for the number of trees and the number of variables sampled at each split were explored to identify the optimal configuration [17]. We defined a grid of values for the number of trees and the number of variables sampled at each split.

We utilized the 5-fold cross-validation to evaluate the RF model's performance across different combinations of hyperparameters. In a 5-fold cross-validation, the dataset is split into 5 subsets, with each subset serving as the validation set once, while the other 4 subsets are used for training. This process helps in assessing the model's generalization ability. The model's performance was fine-tuned by selecting the combination of hyperparameters that minimized the root mean squared error (RMSE), a metric indicating the average difference between observed and predicted values. The RMSE is critical as it directly relates to the model's prediction accuracy, with lower values indicating better performance [18].

To benchmark the performance of the RF model, we also implemented the linear regression (LR) and support vector machine (SVM) models. The LR provides a straightforward baseline, while SVM is known for its effectiveness in high-dimensional spaces. The inclusion of these three algorithms was motivated by their complementary strengths in handling different data characteristics, allowing for a comprehensive comparison of predictive accuracy.

The models were implemented using the Scikit-learn package in Python, a widely used library for machine learning that provides efficient tools for model training, evaluation, and hyperparameter optimization [19].

Following the training process, predictions of breast cancer screening rates were made on a separate testing set. Model accuracy was evaluated using metrics such as $R^2$ and RMSE. $R^2$ represents the proportion of variance in the dependent variable explained by the model, serving as an indicator of goodness-of-fit. RMSE, as previously mentioned, measures the average difference between predicted and observed values, providing insight into the model's prediction error [18].

To interpret the model's predictions, we calculated Shapley Additive Explanations (SHAP) values for each feature. SHAP values provide a detailed understanding of how each feature contributes to the model's predictions [20]. By examining the mean SHAP values, the most influential variables in predicting breast cancer screening rates were identified. For variables with average SHAP values exceeding 0.3, scatterplots were created to explore the direction and magnitude of their effects on screening rates [21].

## Ethical Considerations

The Institutional Review Board at the University of Tennessee Health Science Center determined that this study (24-10240-NHSR) qualifies for Not Human Subjects Research status as it does not involve human subjects as defined by 45 CFR 46.102. The data used in this study were obtained from the publicly available BRFSS dataset provided by the CDC. All study data were aggregated at the census tract level, and no individual-level data were accessed or analyzed, ensuring participant anonymity and compliance with ethical standards.

# Results

## Summary Statistics About Data

Of the 72,337 census tracts nationwide, 49,118 were eligible for inclusion in our analysis, as they had available mammography screening data. The mean mammography screening rate within these census tracts was 77% (SD 3.62) in 2018 and 76.51% (SD 3.71) in 2020. Table 2 provides a detailed overview of summary statistics for all variables considered in our analysis, encompassing the 49,118 included census tracts.

**Table 2.** Summary statistics for all dependent and independent variables for 49,118 census tracts included in the analysis.

| Variables | Missing values, n | Census tracts (N=49,118) |
|---|---|---|
| Mammography rate (2018) (%), mean (SD) | 0 | 77 (3.6) |
| Mammography rate (2020) (%), mean (SD) | 0 | 76.5 (3.7) |
| Location n (%) | | |
|   Rural | 0 | 12,284 (25) |
|   Urban | 0 | 36,834 (75) |
| Population density (per square mile), mean (SD) | 0 | 5,547.38 (13,334.53) |
| Women aged ≥55 years (%), mean (SD) | 0 | 7.9 (4.0) |
| Poverty rate (%), mean (SD) | 49 | 16.3 (12.5) |
| Without health insurance (%), mean (SD) | 35 | 11.47 (7.83) |
| Higher education rate (%), mean (SD) | 4 | 28.4 (18.6) |
| Black population (%), mean (SD) | 1 | 15.2 (23.6) |
| Hispanic population (%), mean (SD) | 1 | 12.7 (18.5) |
| Home value (US $), mean (SD) | 727 | 203,834 (170,438) |
| Social vulnerability index, mean (SD) | 82 | 0.59 (0.28) |
| Primary care shortage, n (%) | | |

| Variables | Missing values, n | Census tracts (N=49,118) |
|---|---|---|
| Yes | 0 | 28,031 (57.1) |
| No | 0 | 21,087 (42.9) |
| Distance to nearest mammography (miles), mean (SD) | 0 | 1.8 (3.2) |
| Number of mammography facilities, mean (SD) | 0 | 18.1 (28.6) |

## Thematic Mapping and Spatial Clustering

Figures 1A and B illustrate the distribution of breast cancer screening rates across the 3143 US counties for the years 2018 and 2020, respectively. Regions in the eastern and northern parts of the country exhibited higher rates of breast cancer screening (>71%), while counties in the central, midwestern, and southern areas displayed comparatively lower rates (<63%). While these visual representations provide valuable insights, further confidence in the findings is derived from statistical and spatial analyses.

Figures 1C and D present the outcomes of Getis-Ord Gi statistics for the clustering of breast cancer screening rates across the United States in 2018 and 2020, respectively. The red areas (hotspots) on these maps represent spatial clusters characterized by high mammography rates, indicating that the screening rates and their neighboring values significantly surpass those in other regions. Conversely, the blue areas denote coldspots, representing spatial clusters with lower screening rates. The similarity in patterns between the two time points underscores the reliability of the observations and strengthens the robustness of the identified spatial clusters. The map also reveals certain disparities. For instance, counties along the western borders, such as California, experienced a decline in mammography rates from 2018 to 2020. Similarly, regions in Indiana, Texas, and Arkansas saw decreased rates of breast cancer screening during this period. Conversely, parts of Illinois and Louisiana showed reported mammography rates from 2018 to 2020.

**Figure 1.** Age-adjusted rates of breast cancer screening in US counties for (A) 2018 and (B) 2020. Spatial clusters in (C) 2018 and (D) 2020.

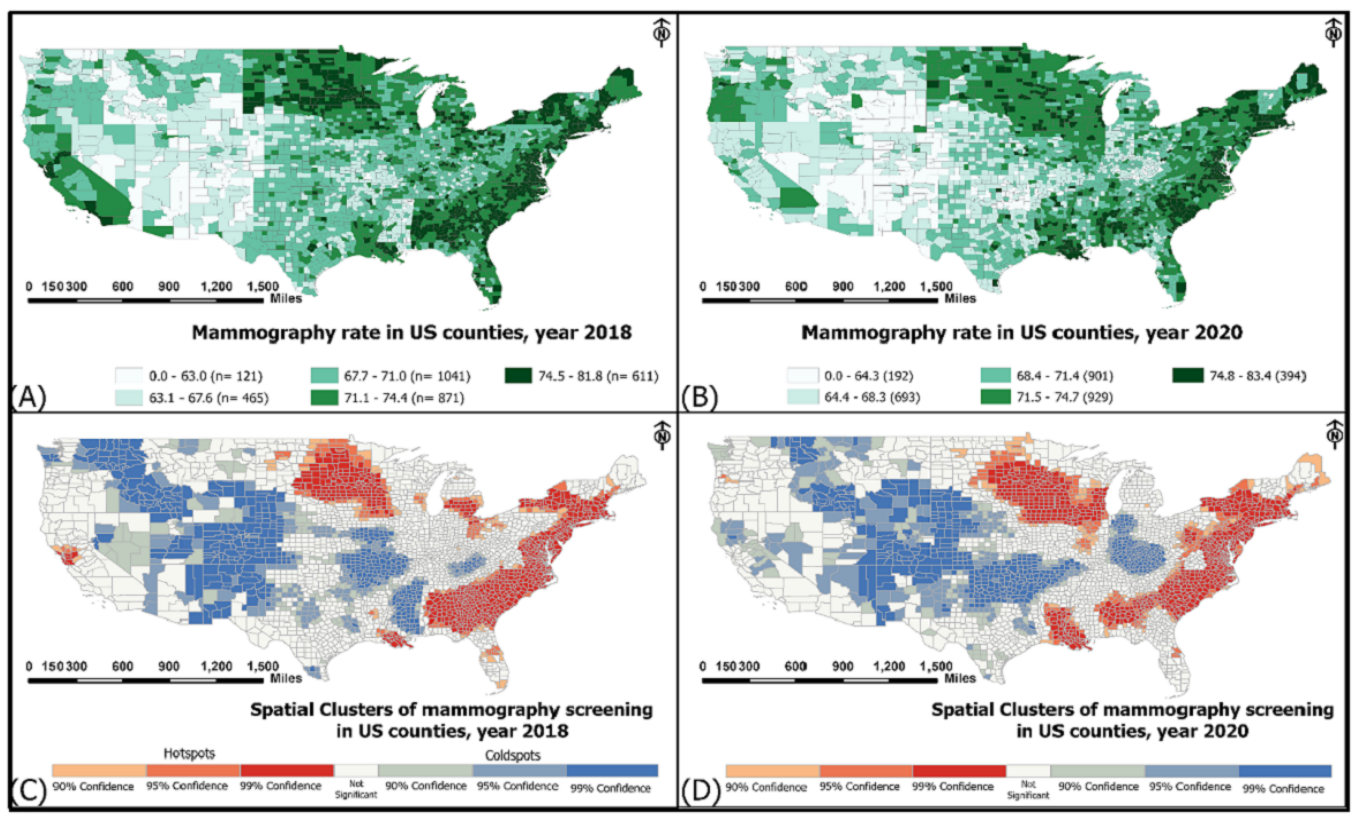

## Machine Learning Analysis

Evaluation of the final RF model, along with the LR and SVM models, based on $R^2$ and RMSE of the testing dataset is presented in Figure 2. The results indicate that the RF model, with an optimal number of trees set to 500 and the number of nodes (m) set to 4, outperforms both LR and SVM. Specifically, the RF model achieved a higher $R^2$ value and a lower RMSE, indicating its superior ability to capture and predict the underlying patterns in the data. This performance underscores the suitability of the RF model for this analysis.

Figure 3 depicts the relative importance of each factor, as determined by SHAP values, at the census tract level in predicting the rate of breast cancer screening across the United States. The mean of SHAP values provides a measure of the overall contribution of each variable to the model's predictions. As evident from Figure 3, the proportion of the Black population is the most important factor, followed by the number of mammography facilities within a 10-mile distance and the higher education rate. For the subsequent analysis, we refined our focus to variables with SHAP values exceeding 0.3 (the top 6 variables) to assess the direction and magnitude of influence that each variable exerts on the prediction of breast cancer screening rates as the variables vary in value.

While assessing variable importance using mean SHAP values offers crucial insights into the most influential factors in predicting breast cancer screening rates, it does not elucidate the direction of their effects on the outcome variable across different variable values. To address this, we generated scatterplots of individual SHAP values for the selected six variables to examine the detailed changes in SHAP values across varying values of these variables. Figure 4 shows that higher proportions of the Black population, higher education levels, an increased number of mammography facilities, and a higher median home value exhibit positive associations with breast cancer screening rates. Conversely, a higher proportion of Hispanic ethnicity and a lack of health insurance demonstrate negative impacts on the screening rates.

**Figure 2.** Comparison of the performance of random forest, linear regression, and SVM models in predicting breast cancer screening rates. RMSE: root mean squared error; SVM: support vector machine.

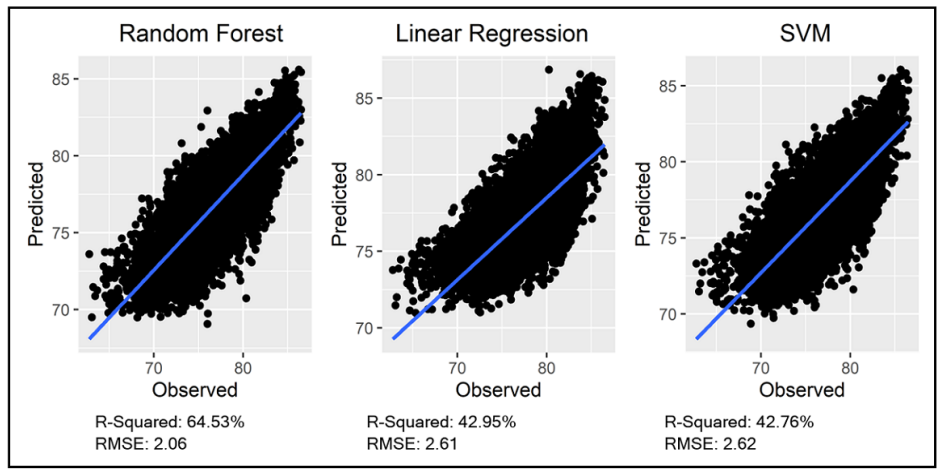

**Figure 3.** SHAP values of each census tract–level factor in predicting the rate of breast cancer screening across the United States. SHAP: Shapley Additive Explanations.

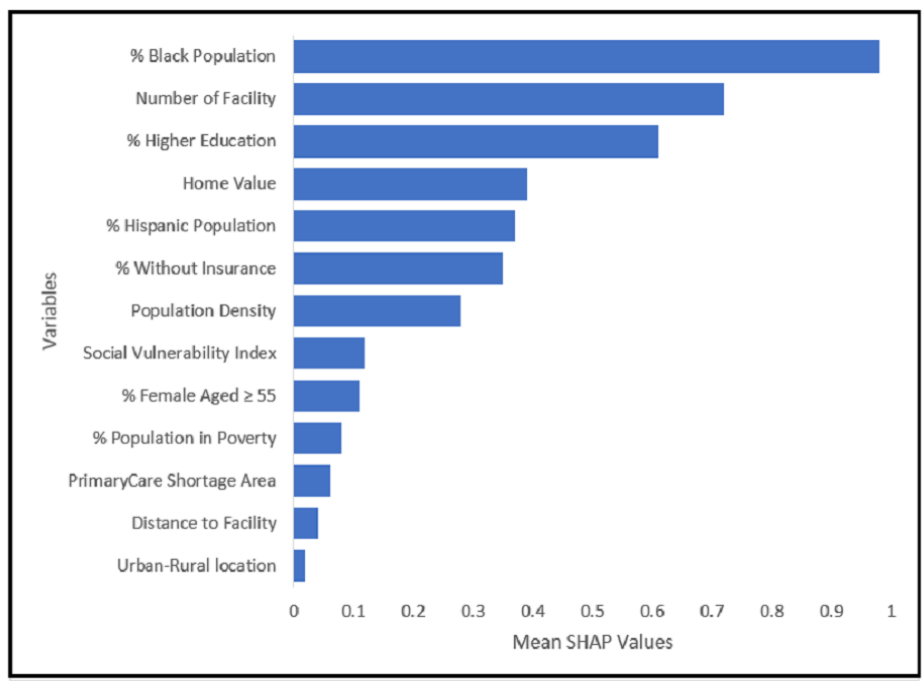

**Figure 4.** Direction of influence illustrated by individual SHAP values on the prediction of breast cancer screening rates for the selected six important variables. SHAP: Shapley additive explanations.

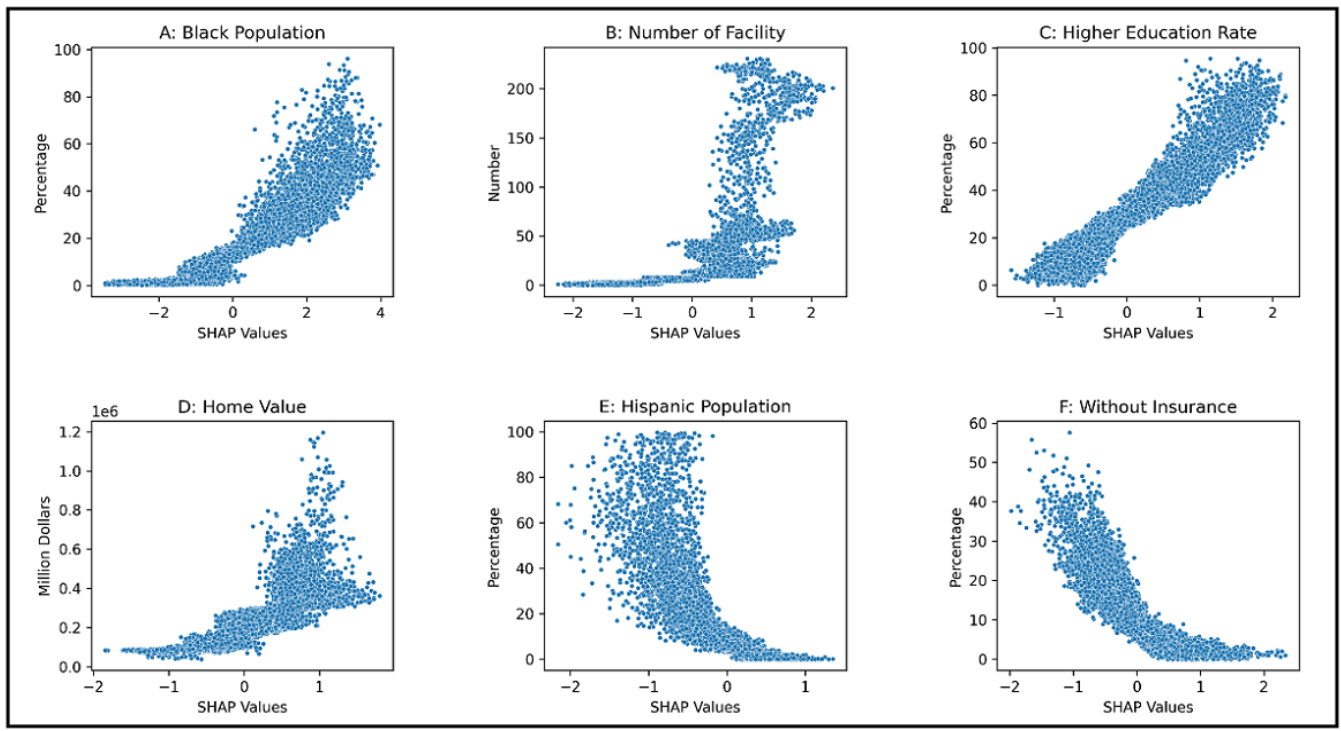

## Discussion

Our research employed a combination of spatial analysis, statistical methods, and machine learning techniques to elucidate the disparities in breast cancer screening across the United States. Spatial patterns revealed clusters of low screening rates, particularly in central, midwestern, and southern regions, contrasted with hotspots of mammography rates, particularly evident along the east coast and in the northern parts of the United States. A predictive model for breast cancer screening rates was developed using the RF algorithm. Meanwhile, key influencing factors for predicting breast cancer screening rates were identified based on the mean SHAP values, including the proportion of the Black population, availability of mammography facilities, and higher education rates.

Spatial clustering identified through Getis-Ord Gi statistics reinforces the observed patterns and underscores their persistence across two distinct time points (2018 and 2020). The consistency of these spatial clusters suggests enduring factors influencing breast cancer screening behavior in specific areas, providing valuable information for policy makers and health care professionals seeking to implement targeted interventions.

It is crucial to acknowledge that the onset of the COVID-19 pandemic has significantly impacted various aspects of human life, including breast cancer screening [21]. The pandemic may have played a role in the decreased mean breast cancer screening rate of breast cancer from 77% in 2018 to 76.51% in 2020. The disruption of routine health care services and the challenges associated with social distancing could have affected mammography screening rates, particularly in urban areas with denser populations. However, it is important to consider that the BRFSS survey focuses on individuals who have undergone breast cancer screening in the last two years. Consequently, women who underwent mammography screening in the year prior to the COVID-19 pandemic could still respond affirmatively. The influence of COVID-19 on the average rate and pattern of breast cancer screening may vary significantly in 2022 and 2023, particularly in cities with higher disease prevalence. Additional investigations are warranted to understand the influence of COVID-19 on changes in breast cancer screening rates across the United States and globally.

Our machine learning analysis that uses an RF model contributes toward understanding the complex interplay of various factors influencing breast cancer screening rates. The RF model with optimal hyperparameters outperformed the LR and SVM models. The ability of RF to capture complex nonlinear relationships and interactions among influencing factors aligns with findings from other population-level studies, highlighting its superiority in predicting population health outcomes [22,23], which confirms our choice of RF as the primary model for our analysis.

While the RF model demonstrated superior performance in predicting breast cancer screening rates, there is a potential risk of overfitting, inherent to ensemble methods [24]. To mitigate this, we implemented cross-validation during the hyperparameter tuning process and evaluated the model's performance on a separate testing dataset to ensure that the RF model maintained its predictive accuracy on unseen data.

It is particularly noteworthy that a higher proportion of the Black population within a census tract was positively associated with increased mammography screening rates. This finding aligns with a 2022 cancer trends progress report, which revealed that 82% of Black women underwent mammography screening, while the screening rate was as low as 74% among other ethnic groups [1]. Possible explanations for this positive association may include the effectiveness of targeted public health interventions and community-based outreach programs specifically designed to increase awareness and accessibility of breast cancer screening in these communities. Additionally, it may reflect a growing awareness and proactive behavior regarding breast cancer prevention among Black women, possibly influenced by public health campaigns and community support networks. Some studies also suggested that when access to health care is equitable, racial and ethnic minorities who are often more aware of their heightened risk, may be more likely to use preventive services like mammography [25]. However, despite the relatively higher mammography screening rates in areas with a larger Black population, it is crucial to underscore that Black women are 40% more likely to die from breast cancer compared to White women [26]. This disparity could be attributed to delays in diagnosis and treatment, particularly when a breast tissue abnormality is identified by mammography [27,28].

Our findings highlighted the importance of the number of available mammography facilities within a 10-mile radius, despite the relatively low SHAP value assigned to the distance to the nearest facility. A plausible explanation is that proximity to a facility may not always be a decisive factor, as various considerations such as affordability and type of insurance can significantly impact facility selection. Moreover, our research revealed that the education rate plays a pivotal role in determining breast cancer screening rates. This finding aligns with prior studies indicating that American women with lower educational attainment are less likely to undergo screening [29]. Educational attainment is closely linked to health literacy [27]; women with lower health literacy have a reduced likelihood of accessing health services, including breast cancer screening [28]. Moreover, women with lower educational attainment might face limited employment opportunities and a lack of jobs that offer access to employee health insurance, leading to a lower likelihood of consulting physicians who recommend mammography.

The variables of home value, rate of the Hispanic population, and rate of the uninsured population exhibited relatively similar and high SHAP values. Areas with higher home values and lower uninsured populations tend to have fewer financial barriers to accessing preventive services. According to existing literature, Hispanic women exhibit lower rates of breast cancer and mortality compared to non-Hispanic Black women and non-Hispanic White women [30]. This disparity could explain the lower screening rate in census tracts with higher proportions of Hispanic population.

Variables analyzed in this study are based on estimates from the CDC PLACES project, which uses a multilevel regression and poststratification approach. This method combines individual-level BRFSS data with demographic data from the US Census to produce reliable estimates at small geographic levels, including census tracts. The multilevel regression and poststratification method has been validated against direct survey data, ensuring that the aggregated rates at the census tract level are both stable and accurate for our analysis [31].

Our study has several limitations. The use of cross-sectional data restricts our capacity to establish causality, underscoring the importance of future research examining temporal changes in breast cancer screening rates. Moreover, as is inherent in all self-reported sample surveys, the BRFSS data may be susceptible to systematic errors stemming from noncoverage, nonresponse, or measurement bias. It is imperative to note that our study was conducted at an aggregate level; therefore, prudence is advised when extrapolating individual-level conclusions. The ecological fallacy, a key concern in population studies, underscores the necessity of avoiding assumptions about individual behaviors based on group-level observations.

This study provides a comprehensive analysis of breast cancer screening disparities in the United States, combining spatial, statistical, and machine learning approaches. The spatial patterns and influential factors identified in this study offer valuable insights for policy makers, health care professionals, and researchers striving to implement targeted interventions to reduce breast cancer screening disparities and improve overall public health outcomes. Ongoing research and targeted interventions are vital for achieving equitable access to breast cancer screening services and ultimately reducing the impact of this significant health issue.

## Acknowledgments

We would like to thank Brianna M White, research coordinator at the Center for Biomedical Informatics, University of Tennessee Health Science Center, for facilitating and assisting in obtaining the institutional review board exemption letter. This study was partially supported by a grant from the Tennessee Department of Health and grant 5 R01MD018766-02 from the National Institutes of Health/National Institute on Minority Health and Health Disparities.

ChatGPT 3.5 was used for grammar checking and language editing of the final manuscript. The core content, analysis, results, and scientific contributions were developed independently by the authors without the use of generative artificial intelligence tools. All content was carefully reviewed and verified by the authors to ensure accuracy.

## Data Availability

The data used in this study were sourced from the Centers for Disease Control and Prevention's Behavioral Risk Factor Surveillance System. These publicly available datasets can be accessed through the Centers for Disease Control and Prevention website, subject to their data use agreement.

**Authors' Contributions**

Conceptualization: SH, ASN, DLS

Data curation: SH, YZ, SWM

Funding acquisition: ASN

Methodology: SH, ASN, DLS

Writing – original draft: SH, FAK

Writing – review and editing: SH, ASN, DLS, YZ, SWM, FAK

Spatial and machine learning analyses: SH, YZ

Supervision: ASN

**Conflicts of Interest**

None declared.

## Abbreviations

**BRFSS:** Behavioral Risk Factor Surveillance System
**CDC:** Centers for Disease Control and Prevention
**LR:** linear regression
**RF:** random forest
**RMSE:** root mean squared error
**SHAP:** Shapley Additive Explanations
**SVM:** support vector machine